\pdfoutput=1

\documentclass[11pt]{article}

\usepackage[]{acl}

\usepackage{times}
\usepackage{latexsym}
\usepackage{todonotes}
\usepackage{booktabs,amssymb,amsmath,multirow}
\usepackage{graphicx}
\usepackage{xcolor}
\usepackage{listings}
\usepackage{subcaption}
\definecolor{codegreen}{rgb}{0,0.6,0}
\definecolor{codegray}{rgb}{0.5,0.5,0.5}
\definecolor{codepurple}{rgb}{0.58,0,0.82}
\definecolor{backcolour}{rgb}{0.95,0.95,0.95}
\lstdefinestyle{mystyle}{
	backgroundcolor=\color{backcolour}
}

\lstdefinelanguage{json}{
	basicstyle=\normalfont\ttfamily,
	numbers=left,
	numberstyle=\scriptsize,
	stepnumber=1,
	numbersep=2pt,
	showstringspaces=false,
	breaklines=true,
	frame=lines,
	backgroundcolor=\color{background},
	literate=
	*{0}{{{\color{numb}0}}}{1}
	{1}{{{\color{numb}1}}}{1}
	{2}{{{\color{numb}2}}}{1}
	{3}{{{\color{numb}3}}}{1}
	{4}{{{\color{numb}4}}}{1}
	{5}{{{\color{numb}5}}}{1}
	{6}{{{\color{numb}6}}}{1}
	{7}{{{\color{numb}7}}}{1}
	{8}{{{\color{numb}8}}}{1}
	{9}{{{\color{numb}9}}}{1}
	{:}{{{\color{punct}{:}}}}{1}
	{,}{{{\color{punct}{,}}}}{1}
	{\{}{{{\color{delim}{\{}}}}{1}
	{\}}{{{\color{delim}{\}}}}}{1}
	{[}{{{\color{delim}{[}}}}{1}
	{]}{{{\color{delim}{]}}}}{1},
}
\colorlet{punct}{red!60!black}
\definecolor{background}{HTML}{EEEEEE}
\definecolor{delim}{RGB}{20,105,176}
\colorlet{numb}{magenta!60!black}

\usepackage[T1]{fontenc}

\usepackage[utf8]{inputenc}

\usepackage{microtype}
\usepackage{graphicx}
\usepackage[framemethod=tikz]{mdframed}
\usepackage{enumitem}
\usepackage{pifont}
\usepackage{makecell}
\newcommand{\smtt}[1]{\mbox{\small\tt #1}}

\usetikzlibrary{shadows}
\newmdenv[
	shadow=true,
	shadowsize=5pt,
	frametitle= MTurk prompt
]{note}

\title{Cross-TOP: Zero-Shot Cross-Schema Task-Oriented Parsing}

\author{Melanie Rubino\\
  Amazon Alexa AI \\
  New York, USA \\
  \texttt{rubinome@amazon.com} \\\And
  Nicolas Guenon des Mesnards \\
  Amazon Alexa AI \\
	New York, USA \\
  \texttt{mesnarn@amazon.com} \\ \And
Uday Shah \\
Amazon Alexa AI \\
New York, USA \\
\texttt{shahuda@amazon.com} \\
\AND
Nanjiang Jiang \\
Department of Linguistics\\
The Ohio State University \\ 
\texttt{jiang.1879@osu.edu} \\\And
Weiqi Sun \\
Amazon Alexa AI \\
New York, USA \\
\texttt{weiqisun@amazon.com} \\\And
Konstantine Arkoudas \\
Amazon Alexa AI \\
New York, USA \\
\texttt{arkoudk@amazon.com} \\
} 

\begin{document}
\maketitle

\begin{abstract}
Deep learning methods  have enabled task-oriented semantic parsing of increasingly complex utterances. However, a single model is still typically trained and deployed for each task separately, requiring labeled training data for each, which makes it challenging to support new tasks, even within a single business vertical (e.g., food-ordering or travel booking). In this paper we describe Cross-TOP (Cross-Schema Task-Oriented Parsing), a zero-shot method for complex semantic parsing in a given vertical. By  leveraging the fact that user requests from the same vertical share lexical and semantic similarities, a single cross-schema parser is trained to service an arbitrary number of tasks, seen or unseen, within a vertical. We show that Cross-TOP can achieve high accuracy on a previously unseen task without requiring any additional training data, thereby providing a scalable way to bootstrap semantic parsers for new tasks. As part of this work we release the FoodOrdering dataset, a task-oriented parsing dataset in the food-ordering vertical, with utterances and annotations derived from five schemas, each from a different restaurant menu.
\end{abstract}

\section{Introduction}

Propelled by deep learning, task-oriented parsing has  made significant strides, moving away from flat intents and slots towards more complex tree-based semantics 
that can represent compositional meaning  structures  \cite{gupta2018semantic,aghajanyan-etal-2020-conversational,rongali2020don,mansimov2021semantic}. 
However, most semantic parsing systems remain task-specific: they can only produce representations with the set of intents and slots seen during training.
To support multiple tasks, this approach requires collecting data, training, and maintaining a model for each task separately.
This is costly when multiple tasks need to be supported, as is usually the case for digital voice assistants such as Alexa and Google Assistant, which may need to support hundreds or thousands of different tasks in a given business vertical (e.g., restaurants in the food-ordering vertical, hotels in the travel vertical, and so on).

\begin{figure}[h!]
	\hspace*{-15pt}
    \centering
       \includegraphics[width=0.54\textwidth]{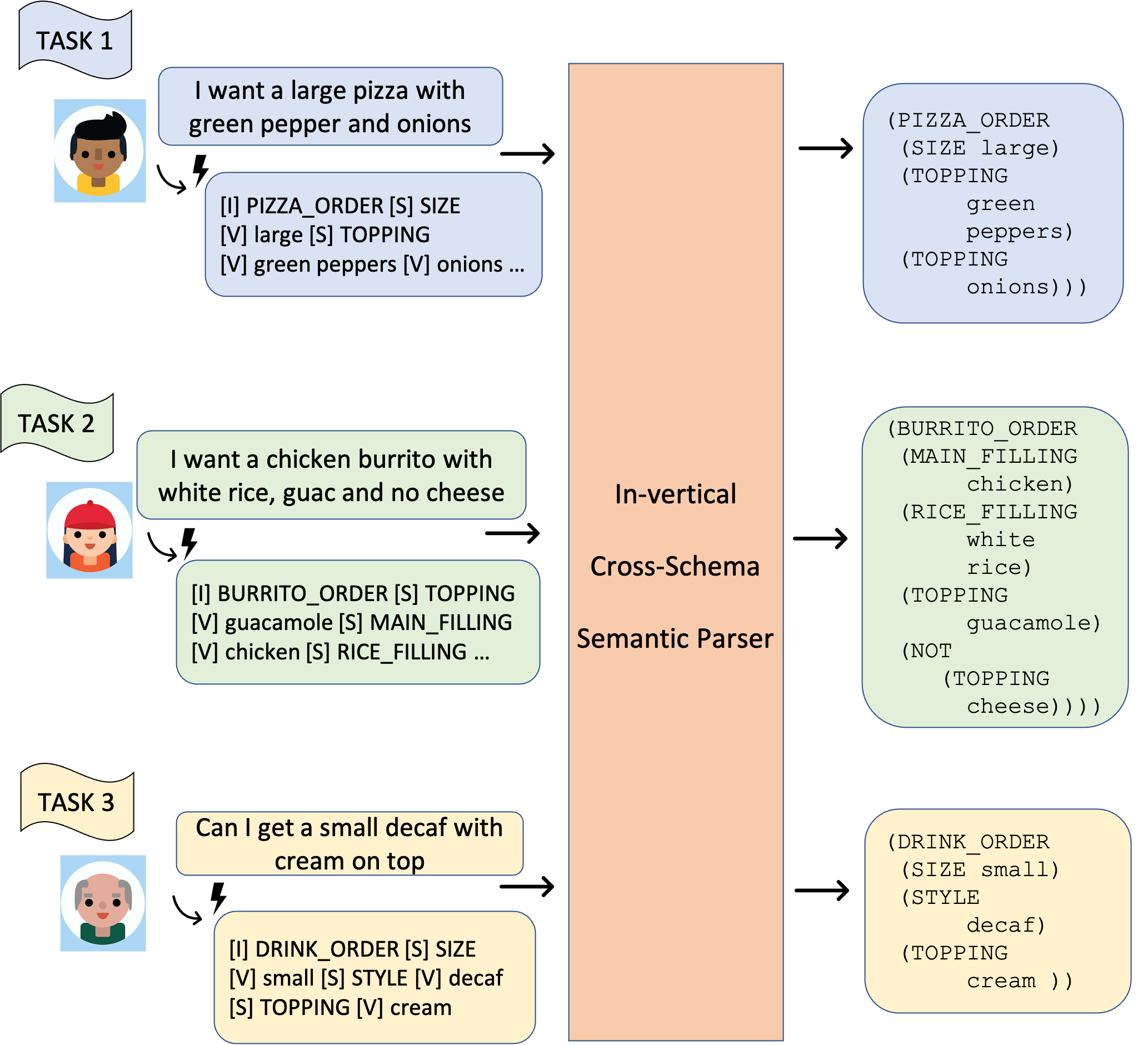}
    \caption{The Cross-TOP parser processes utterances from multiple tasks with different schemas. The lightning bolts represent fuzzy matching, which is used to append schema elements to the input (cf. Section \ref{sec:model}).}
    \label{fig:intro_example}
\end{figure}

In this paper we present Cross-TOP, a method for building a single semantic parsing model that can support an arbitrary number of tasks in a given vertical. User requests pertaining to the same vertical have lexical and semantic similarities; their main differences lie in their unique schemas. 
In the food-ordering domain, for example, a customer may request a main dish with various options and possibly a drink and a side. However, 
depending on the specific restaurant menu, the output semantic representations can differ greatly; see Figure~\ref{fig:intro_example}.

Cross-TOP makes use of a powerful pretrained transformer-based encoder-decoder
language model, with schema-specific context added to the input along with the utterance.  In this way, the model learns to generate parses 
for a new, unseen task, by attending to the schema in the input rather than by needing to see it during training.
We show that this approach is quite effective and provides a quick solution to the practical problem of bootstrapping semantic parsers 
for new tasks within a vertical, using a single model in production.

The parser is trained on a number of initial tasks, where each task has some training data available. Moreover, we assume that every task has a unique {\em schema}. 
That schema consists of all possible intents and slots for the task at hand; both intents and slots can be arbitrarily nested (compositional). 
For every slot $S$, the schema also includes natural-language phrases for the various values of $S$. All schemas for the five tasks in our dataset can be found in Appendix \ref{appendix:skillSchemas}. Cross-TOP uses constrained decoding to ensure that it generates well-formed parses that can be resolved to executable representations that can be directly used by the back end.

Most zero-shot cross-schema semantic parsing work has been in the context of the Text-to-SQL task~\cite{zhong-etal-2020-grounded, lin-etal-2020-bridging, wang-etal-2020-rat, rubin-berant-2021-smbop-semi, DBLP:journals/corr/abs-2009-13845, gan2021natural}.
Cross-schema task-oriented parsing introduces its own challenges.
In SQL, the schemas are database schemas, and the parser is trained on some initial databases and then evaluated on another database.
There is a lot of invariant structure across different tasks in the output space (since output sequences are always SQL queries), as well as common patterns in how SQL structures tend to align with natural language. However, for the schemas defined in task-oriented parsing 
for the food-ordering domain, the only invariant structures are the parentheses and some lexical overlap among the intents and slots.
Therefore, cross-schema parsing in general is more challenging for task-oriented parsing. However, restricting the scope to a given vertical imposes more common structure that can prove helpful.

To evaluate our methodology, we focus on tasks in the food-ordering domain, where each task contains examples from a
restaurant with the schema generated from its menu.
Our main contributions are as follows:
\begin{itemize}
    \item We present a new technique for zero-shot intra-vertical cross-schema semantic parsing that jointly encodes utterance tokens and schema elements.
    \item We release a new task-oriented parsing dataset for food ordering to evaluate similar efforts. The FoodOrdering dataset includes examples from five restaurants,
    totaling close to 30,000 synthetically-generated training examples and 963 human-generated test utterances with labels.
    \item We show that our method achieves up to 73\% exact match accuracy on a previously unseen ordering task, proving the method's viability for effortlessly handling a new task.
\end{itemize}

\section{Model}\label{sec:model}
Our method trains a single schema-aware model to serve multiple tasks and bootstrap new ones from the same business vertical in a zero-shot setting.  It leverages the transfer learning capabilities of a transformer-based pretrained encoder-decoder language model.

\paragraph{Terminology} Each \emph{task} is defined by a unique \emph{schema} consisting of \emph{intents}, \emph{slots} belonging to those intents, and \emph{catalogs} enumerating the possible \emph{slot values} for each slot. For example, in the pizza-ordering task the \smtt{TOPPING} slot belongs to the \smtt{PIZZAORDER} intent, and values for this slot could be \smtt{mushrooms}, \smtt{pepperoni}, and so on. In our predefined catalogs, multiple slot values could refer to the same \emph{slot value entity}, for example \smtt{peppers} and \smtt{green peppers} can both be mapped to \smtt{TOPPING\_PEPPERS}---or perhaps \smtt{TOPPING\_35}---in the back end. Cross-TOP predicts parse trees that contain slot values, which are then entity-resolved into those unique back-end identifiers through this many-to-1 mapping.

A task schema can optionally define \emph{invocation keywords} for each intent, to identify how these are expressed in natural language, for example \{\emph{drink}, \emph{drinks}\} for a \smtt{DRINK\_ORDER} intent. This is used for augmenting model input with \emph{fuzzy-matched} schema elements later on.
Fuzzy string-matching algorithms compute lexical similarity between strings. If some schema elements have a significant overlap with certain utterance tokens, then there is a ``match'' and that schema element will be appended to the input utterance before encoding.

\paragraph{Model Inputs} As just mentioned, to achieve zero-shot cross-schema parsing, we
append fuzzy-matched schema elements to input utterances.
 Given an utterance \texttt{u}, assume our fuzzy-matching process (described later) determined that the intents \texttt{i$_1$} and \texttt{i$_2$} are present in the request, with slot/slot-values \texttt{s$_{1,1}$}/\texttt{v$_{1,1}$} for \texttt{i$_1$}, as well as \texttt{s$_{2,1}$}/\texttt{v$_{2,1}$} and \texttt{s$_{2,2}$}/\texttt{v$_{2,2}$} for \texttt{i$_2$}\footnote{There can be more than one slot value \texttt{v} identified for the same slot \texttt{s}, in which case the input will be of the form:           \texttt{$\quad$ u [I] \texttt{i$_1$} [S] s$_{1,1}$ [V] v$_{1,1,1}$ [V] v$_{1,1,2}$ $\ldots$}}. The input to Cross-TOP is then serialized into the following format:


\noindent\texttt{u [I] i$_1$ [S] s$_{1,1}$ [V] v$_{1,1}$  [I] i$_2$ [S] s$_{2,1}$  [V] v$_{2,1}$ [S] s$_{2,2}$ [V] v$_{2,2}$ }

\begin{figure}[hbt!]
	\hspace*{-10pt}
       \includegraphics[width=0.5\textwidth]{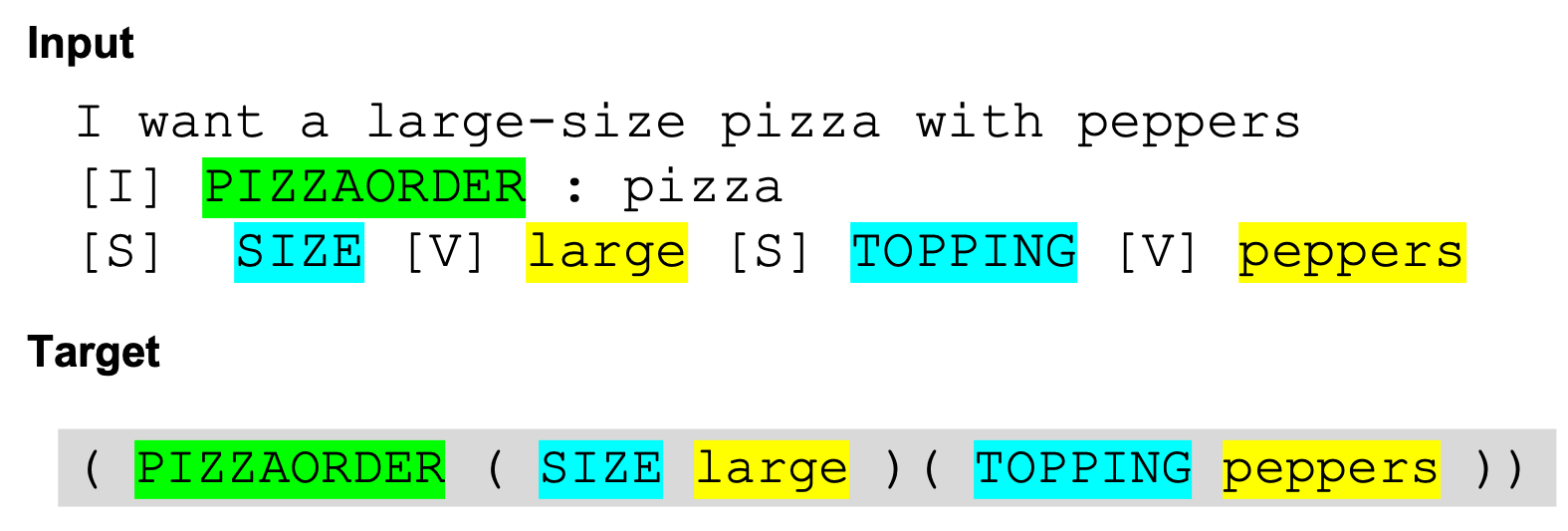}
    \caption{Cross-TOP is trained to attend to input utterances augmented with fuzzy-matched schema elements.}
    \label{fig:attend}
\end{figure}

\smallskip

\noindent where markers \texttt{[I]}, \texttt{[S]}, \texttt{[V]} indicate that the following tokens are intents, slots and slot values, respectively. An example is given in Figure \ref{fig:attend}. 

Our format is inspired from BRIDGE \citep{lin-etal-2020-bridging}, but instead of table/column/column-value in a database schema, task-oriented parsing schemas uses intent/slot/slot-value. While the longer input sequences increase the computation required for inference, the latency impact is mitigated by the parallelizability of the transformer architecture.

\paragraph{Model Outputs}
The model is trained to generate a linearized parse tree similar to the target shown in Figure \ref{fig:attend}, which is reminiscent of the 
TOP {\em decoupled } notation \cite{aghajanyan-etal-2020-conversational}. TOP  decoupled is itself derived from the TOP notation \cite{gupta2018semantic} by removing tokens that are not direct children of slot nodes. Unlike TOP decoupled, leaf nodes in our output semantics are not tokens copied from the source utterance, but instead must be valid \emph{slot values} belonging to the task's catalogs.  As exemplified in Figure \ref{fig:attend}, the fuzzy-matched slot value for the utterance segment \smtt{large-size} is the catalog entry \smtt{large}. It can happen that utterance token and catalog value are identical, as is the case for \smtt{peppers} here. By predicting slot values instead of unresolved utterance tokens, Cross-TOP jointly learns to perform semantic parsing and entity resolution, thus eliminating the need to train and maintain a separate entity resolution system for every new task.

\paragraph{Fuzzy-Matching Details}\label{subsec:fuzzy} The viability of our schema-aware encoding depends on our ability to extract the proper schema elements. We leverage the fuzzy-matching method from the BRIDGE codebase\footnote{\url{https://github.com/salesforce/TabularSemanticParsing}} and compute lexical similarity scores between an input utterance and every slot  value.\footnote{This works in a vertical with small catalogs, such as restaurant menus. To make it scale to much larger catalogs, one could use sub-linear fuzzy string-matching algorithms and offline parallel processing.} If multiple slot values representing the same entity match the utterance, we pick the one with the higher similarity score. Slots are added to the input if at least one of their slot values was added.\footnote{Slots that are parents of other slots are also provided with catalog entries to allow fuzzy matching. For example, a \smtt{NOT} slot for negation will use \{\emph{with no}, \emph{without}, \emph{hold the} $\ldots$\}.} Intents are added to the input if at least one of their slots is added.\footnote{A slot shared across two intents will trigger both their inclusion, but experiments indicate that the neural parser can learn to discard such false detection.}

In addition, if any of the predefined intent invocation keywords (cf. \textbf{Terminology}) fuzzy-match the utterance, then that intent is added along with the fuzzy-matched keyword, for example adding \smtt{[I] PIZZAORDER : pizza} instead of simply \smtt{[I] PIZZAORDER}. Given that intent names can be arbitrary and carry little semantic content, this design helps the pretrained language model by bridging the gap between natural language and back-end executable representations.

\paragraph{Constrained Decoding} The target parses contain only schema elements and parentheses. Cross-TOP leverages constrained decoding at inference time to generate valid catalog values and parses according to the schema. For example, the string \smtt{(DRINK\_ORDER (SIZE coke))} is not valid, as the slot value \smtt{coke} is not a catalog entry for the slot \smtt{SIZE}. In this work we also implement a parentheses-balancing constraint, as well as a set of valid next-token constraints, where each
vocabulary subword has a corresponding entry in a dictionary mapping it to a list of valid subwords that may follow it. The content of such a dictionary is task-specific but is built programmatically from the task schema. The detailed constraints are provided in Appendix~\ref{appendix:constrained}. Section~\ref{section:results} quantifies the benefits of constrained decoding in the zero-shot setting.


\section{The FoodOrdering Dataset}
We release a dataset for cross-schema zero-shot task-oriented parsing: the FoodOrdering dataset,\footnote{\url{https://github.com/amazon-research/food-ordering-semantic-parsing-dataset}}  comprising five food-ordering tasks for five fictitious restaurants: \textsc{pizza}, \textsc{sub}, \textsc{burrito}, \textsc{burger} and \textsc{coffee}.

\paragraph{Dataset Construction} To gauge zero-shot capabilities, only three out of five  tasks come with training data. For \textsc{sub} and \textsc{burrito}, the training portion of the data was synthetically generated by sampling around 50 human-designed templates for which slot values are themselves sampled from predefined catalogs. The catalogs and templates are released along with the dataset, but a couple of examples are given in Table \ref{table:synthEx}. We generated up to 10,000 unique pairs of natural language and target parses. For \textsc{pizza} we randomly sampled 10,000 utterances out of the 2.5M provided by \citet{pizzaDataset}. All five tasks have evaluation data generated by humans and collected through Mechanical Turk; see Appendix \ref{appendix:mturk} for details. MTurk workers generated natural language orders, which were then annotated internally. More examples can be found in Appendix \ref{appendix:dataExamples}.

\paragraph{Dataset Statistics}  All tasks follow a common structure of intents and  slots, but each task has a different number of intents, slots and slot values. 
In Table \ref{table:exrres}, the \#SltValEntities column does not count the total number of slot values, but rather the total number of slot value entities, which are resolved slot values (cf. \textbf{Terminology}). \textsc{burrito} has 7 distinct intents while \textsc{coffee} is a single-intent task. The design differences between the task schemas reflect a real-world setting: each restaurant comes with its own preexisting back end that dictates the design and contents of the corresponding schema.  On average there are 1.7 intents and  6.2 slots per utterance, and an average depth\footnote{Queries are by design multi-intent, hence implicitly rooted in a parent \smtt{ORDER} node, which is factored in the computation of depth.} of 3.4.  Detailed numbers are provided in Table~\ref{table:uttStats} of Appendix~\ref{appendix:dataExamples}.

\begin{center}
	\begin{table*}[!hbt]

			\centering
			\begin{center}
				\footnotesize
				\begin{tabular}[c]{lcccccl}
					\hline
					Dataset 					& \#Train/Synthetic  &  \#Eval & \#Int & \#Slt & \#SltValEntities &  Example utterance \\
					\hline
					\textsc{Pizza}                      & 10,000 & 348 & 2 &  10 &  166 & \makecell[l]{ \textit{"Can i get one large pie with no} \\ \textit{cheese and a coke."} } \\
					\textsc{Burrito}                      & 9,982 & 191 & 7 &  11 &  34 & \makecell[l]{ \\ \textit{"One carnitas quesadilla with} \\  \textit{white rice and black beans."}} \\
					\textsc{Sub}                       & 10,000 & 161 & 3 &  8 & 62 &  \makecell[l]{ \\ \textit{"Get me a cold cut combo with} \\ \textit{mayo and extra pickles."}} \\
					\textsc{Burger}                      & 0* & 161 & 3 &  9 & 44  &  \makecell[l]{ \\ \textit{"A vegan burger with onions and a} \\ \textit{side of sweet potato fries."}} \\
					\textsc{Coffee}                      & 0* & 104 &1 &  9 &  43 & \makecell[l]{ \\ \textit{"One regular latte cinnamon iced with} \\ \textit{one extra espresso shot."}} \\
					\hline
				\end{tabular}
			\end{center}
			\caption{FoodOrdering dataset statistics: sizes of training and evaluation sets, as well as numbers of intents, slots, and resolved slot value entities defined in each task's schema. *\textsc{burger} and \textsc{coffee} have no training data, as they are used to evaluate zero-shot learning.}
			\label{table:exrres}
	\end{table*}
\end{center}

%

\paragraph{Task Schemas} Each task has a unique schema, but all schemas are governed by similar rules: only slot nodes can be children of intent nodes, and there is no limit on the number of intents per utterance nor slots per intent. Slot nodes can be parents either of slot values or of other slots. \smtt{NOT} is an example of a generic (task-agnostic) slot that allows us to negate any slot that admits negation, such as \smtt{TOPPING}. Refer to Appendix~\ref{appendix:skillSchemas} for the details of the five schemas.

\section{Experimental Setup}

Our experimental setup reflects the practical scenario of having to scale a technology to service multiple applications under constrained production resources. We consider a single model to serve all tasks, so we train with synthetic data for only three of the tasks (\textsc{pizza}, \textsc{burrito} and \textsc{sub}), and test zero-shot generalization on two unseen tasks (\textsc{burger} and \textsc{coffee}).

\paragraph{Training Details}
In this work we use BART-Large \cite{lewis-etal-2020-bart}, a transformer-based pretrained encoder-decoder language model.
We fine-tune the publicly available 24-layer BART-Large checkpoint\footnote{\url{https://huggingface.co/facebook/bart-large}} totaling 406M parameters, using the \texttt{transformers} codebase. We expand the target vocabulary by adding special tokens for input markers \texttt{[I]}, \texttt{[S]} and \texttt{[V]}.
The training dataset was created by concatenating synthetic data from the three training tasks.
Models are trained for 50 epochs with early stopping patience of 4,
using cross-entropy sequence loss and the \texttt{AdamW} optimizer. We use the human-generated data of the three training tasks as our development set for early stopping and hyperparameter tuning. Hyperparameter tuning is described in Appendix~\ref{appendix:comput}. Our best model uses a batch size of 16, learning rate \texttt{1e-05} and linear learning rate scheduler with warm-up ratio of 0.1. The hyperparameter \texttt{no\_repeat\_ngram\_size} was disabled by setting it to 0.

\paragraph{Evaluation Details} We use Unordered Exact match accuracy (Unordered EM) to measure performance. It checks for an exact match between the golden and predicted trees, where sibling order does not matter.
The golden parse trees are executable representations (ready for consumption by an appropriate back end) that contain resolved entity names instead of slot values identified by utterance segments. These entities are fully determined by the many-to-1 mapping mentioned in Section~\ref{sec:model}. Validation performance is computed on the aggregated validation sets for the three training tasks. Test performance is reported for tasks individually. We used a beam size of 6 for validation and testing.

\paragraph{Pre-Processing and Post-Processing}
When appending the schema elements to the input utterance we do not include the slot/slot-value pair \smtt{NUMBER}, \smtt{1} from the fuzzy matching process if it's the only quantity matched.\footnote{Note that we do keep those slot/slot-value pairs for quantities larger than 1.} This choice was made after observing that the slot values \smtt{a/an} can easily trigger false positives in fuzzy matching. For example, in the utterance \emph{an order of two sprites}, the numeric quantity to extract is \emph{two}, but the token \emph{an} would trigger an extra unnecessary match. At inference time, if no \smtt{NUMBER} was generated for an intent, we add back \smtt{(NUMBER 1)} as a default to the predicted parse tree. Before computing unordered EM scores, all slot values are resolved into the appropriate entity names using the many-to-1 mapping mentioned earlier.

\section{Results and Analysis}\label{section:results}
In the zero-shot setting, Cross-TOP achieves 73\% unordered EM on \textsc{burger} and 55\% on \textsc{coffee} (Table \ref{table:main_results}). The rest of this section presents an analysis of our results. 

\paragraph{Schema-aware encoding enables zero-shot transfer learning.} The main strength of Cross-TOP is  training and maintaining a single model that can serve multiple tasks within the same business vertical, and bootstrapping new tasks without retraining. The zero-shot results in Table \ref{table:main_results} support the claim that joint learning over utterance tokens and matched schema elements achieves this objective. For completeness, we show that the zero-shot ability does not simply come from the conjunction of constrained decoding and BART's extensive pretraining: we perform an ablation exercise where the input to BART-Large contains no schema information at all, but constrained decoding is enabled. As can be seen in the second row of Table \ref{table:main_results}, accuracy drops precipitously, by 46 and 23 absolute points for \textsc{burger} and \textsc{coffee}, respectively.
A manual analysis of the predictions shows that in the overwhelming majority of cases, this model only generates  intents and slots that it has seen before in training 
and thus fails to correctly parse utterances that have unseen intents/slots. On a subset of 108 \textsc{burger} utterances with at least one intent unseen in training, the schema-oblivious approach only gets 4\% unordered EM, compared to 64\% for Cross-TOP.

\paragraph{Schema-aware decoding ensures proper executable parses.} Schema-aware constrained decoding ensures that Cross-TOP generates fully executable parse trees. Without this component, performance drops by 20 absolute points, as shown in the third row of Table~\ref{table:main_results}. By looking at 15 predicted utterances where the output predictions change by removing constrained decoding, we found that 93.3\% of \textsc{burger} utterances and 60\% of \textsc{coffee} utterances contained at least one invalid slot/slot-value combination. While using constrained decoding on these utterances is guaranteed to rule out invalid combinations, this does not ensure that the result will be correct. However, on inspection we found that constrained decoding transforms  60\% of \textsc{burger} and 33.3\% of \textsc{coffee} mismatched utterances to be completely correct. Table~\ref{table:constrained_decoding_helps_examples} in Appendix~\ref{appendix:constrained} illustrates how constrained decoding can help with specific examples.

\paragraph{Cross-TOP improves as more training tasks are added.}
While our main result shows that training with only few tasks allows zero-shot transfer to new tasks with no retraining, a realistic production scenario would be to periodically retrain the model by incorporating new training data. To quantify the benefits of adding more tasks, we compare training Cross-TOP using one, two or three tasks. The results for training on one task are an average over three models, one trained on \textsc{pizza} only, one on \textsc{burrito} and one on \textsc{sub}.  Likewise, results for training on two tasks are an average of three models, one trained on
\textsc{pizza+burrito}, one on \textsc{burrito+sub} and the other on \textsc{pizza+sub}. As shown in Table~\ref{table:main_results}, going from 1 to 2 tasks doubles performance for \textsc{burger}, and using 3 tasks almost triples the performance for both test tasks, confirming that the model learns general patterns that govern all schemas in the food-ordering vertical. 

\paragraph{Dependency on fuzzy matching}
Cross-TOP relies on the quality of the fuzzy-matching process that determines which schema elements are encoded along with the utterance tokens. It can be challenging to recover from a fuzzy matching failure that ends up omitting a slot value from the input. In \textsc{burger}, such failures account for only 1\% of all test utterances. In \textsc{coffee} that phenomenon is more prominent, with 7\% of test utterances presenting at least one missing element from the fuzzy-matched schema.
These limitations can be addressed by making the fuzzy-matching algorithm more robust and/or by adding unrecognized slot values as extra entries in the slot's catalog. The latter option is appealing, as it involves no model retraining, but does not suffice, as there is no obvious way to automate it. We upper-bound the impact of any candidate fix by providing an oracle schema for all utterances, and observe in the last row of Table~\ref{table:main_results} that it brings an absolute improvement of 2 absolute point in \textsc{burger} and 5 absolute points in \textsc{coffee}.



\begin{center}
    \begin{table}[!hbt]
    \begin{center}
        \small
        \hspace{-0.4cm}
        \begin{tabular}[c]{lll}
            \toprule
                                & Burger & Coffee \\
              \midrule
             Cross-TOP       & 73.3 $\pm$ 3.6  & 54.8 $\pm$ 5.7 \\
             \midrule
	          $\,$ w/o schema-augmented input  $\;$   & 26.5 $\pm$ 1.5  & 31.7 $\pm$ 1.0 \\
	          $\,$ w/o constrained decoding  $\;$    & 53.0 $\pm$ 4.2 & 33.3 $\pm$ 7.2 \\
        	 $\,$ training only on 1 task $\;$  & 25.4 $\pm$ 1.6 & 19.9 $\pm$ 3.3 \\     
	          $\,$ training only on 2 tasks $\;$  & 52.4 $\pm$ 2.7 & 34.0 $\pm$ 3.5 \\           
	          $\,$ w/ oracle schema $\;$  & 75.6 $\pm$ 4.3 & 59.4 $\pm$ 7.1 \\               
            \bottomrule
        \end{tabular}
    \end{center}
    \caption{Cross-TOP zero-shot unordered EM accuracy, averaged over 3 seeds, along with various ablations. The $\pm$ signs indicate the standard error across seeds.}
    \label{table:main_results}
    \end{table}
\end{center}


\section{Related Work} \label{sec:related}

\paragraph{Slot Filling}
Traditionally, task-oriented parsing for flat intents and slots
has been framed as a combination of intent classification and slot labeling~\cite{7846294}, possibly
with an additional domain classification component. 
Several authors have addressed zero-shot solutions in this field. 
QASF~\cite{du-etal-2021-qa} is a QA-driven approach that extracts slot-filler spans from utterances using a question-answering model.
Both~\citet{bapna2017zeroshot} and~\citet{siddique2021linguisticallyenriched} tag words with slots using slot descriptions
and context-aware representations of the utterance.
These solutions don't apply to structured (compositional) semantic representations, or to multiple intents in a single utterance, both of which are handled by Cross-TOP.

\paragraph{Task-Oriented Parsing}
In the more general area of task-oriented parsing, where hierarchical representations are featured, the authors are not aware of
other zero-shot cross-schema work. There is some work in the few-shot setting~\cite{chen-etal-2020-low}, where data from multiple domains is used during an additional
stage of fine-tuning combined with meta-learning.

\paragraph{Text-to-SQL}
Some of the most relevant related zero-shot work is in text-to-SQL semantic parsing.  In this area, a challenging
dataset, SPIDER~\cite{yu-etal-2018-spider}, is the most common dataset used to test zero-shot solutions. 
The GAZP~\cite{zhong-etal-2020-grounded} method generates synthetic training data for the new schema environment and
requires a retraining of the neural parser, not making it as convenient of a zero-shot method. RAT-SQL~\cite{wang-etal-2020-rat} moves away from needing to
retrain the parser, and focuses on jointly encoding the schema and utterance tokens.
BRIDGE \citep{lin-etal-2020-bridging} is the main inspiration for our work, as it  encodes the utterance and schema together, and augments the input with anchor texts, which
are database values from tables, designed to better bridge utterance tokens to database tables, columns and values. 
Another notable contribution intended to bridge the gap between natural language and machine-executable representations is the work of \citet{gan2021natural}, which leverages an intermediate representation to go from text to SQL. 



\section{Conclusion}

We presented Cross-TOP, a zero-shot method for cross-schema task-oriented parsing that eliminates the need to retrain and maintain a new model for every new task in a business vertical. We released a new dataset illustrative of five real-world applications in the food-ordering vertical. We showed that Cross-TOP reaches up to 73\% EM accuracy in zero-shot transfer, making it a viable technique for quickly bootstrapping a parser for a new task.



Future work could further enrich the joint encoding of utterances and task schemas, while an additional thread of work could study how to best leverage limited annotated data that may be available for a new task.

\section*{Acknowledgments}
	We would like to thank Beiye Liu, Emre Barut, Ryan Gabbard, and anonymous reviewers for providing valuable feedback on this work.

\bibliography{custom}
\bibliographystyle{acl_natbib}

\newpage
\clearpage

\appendix
\section*{Appendix}

\section{Mechanical Turk Instructions}\label{appendix:mturk}
\begin{figure*}[b]
	\centering
	\begin{subfigure}[b]{0.475\textwidth}
	\includegraphics[width=\textwidth]{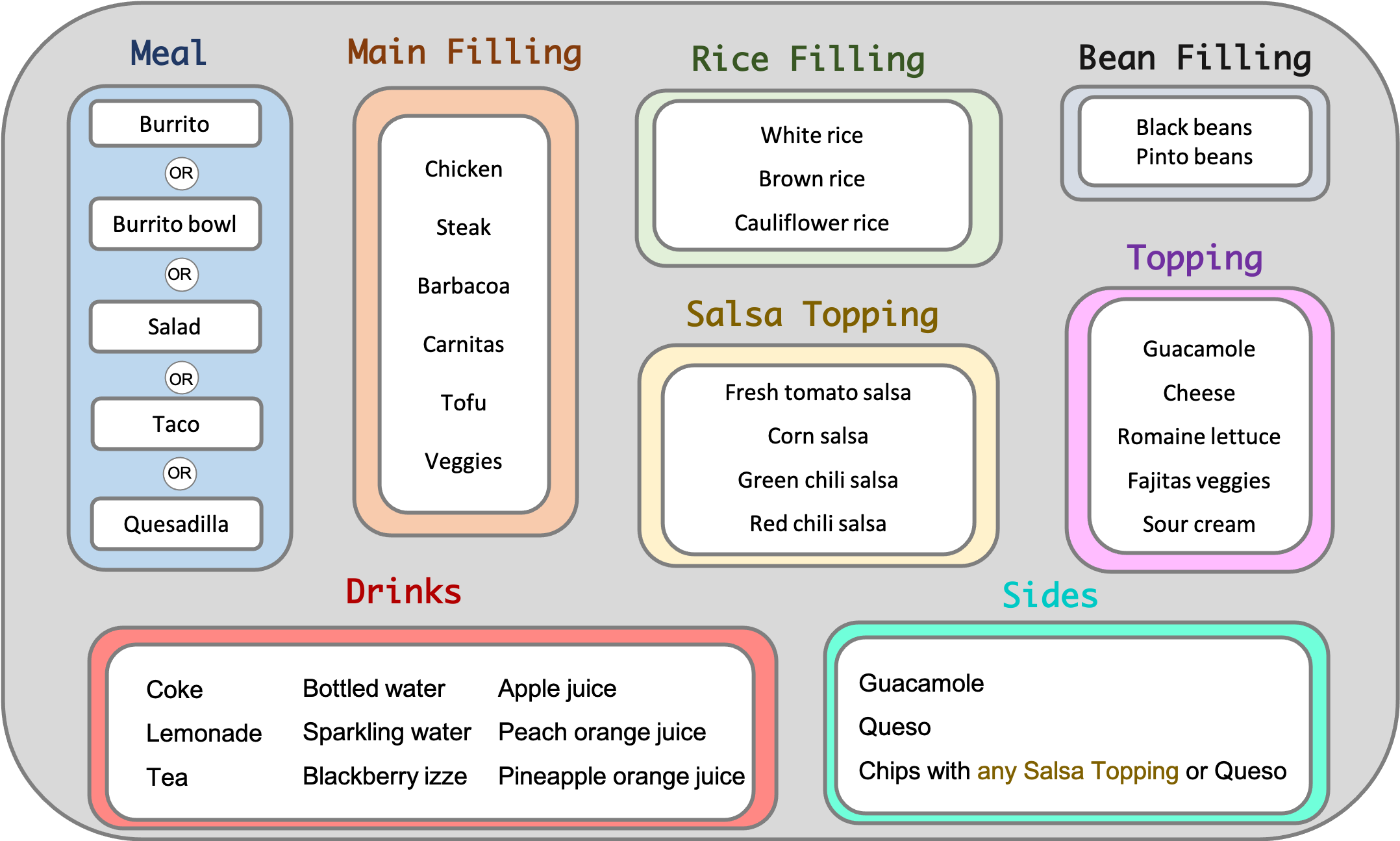}
	\end{subfigure}
	\hfill
	\begin{subfigure}[b]{0.475\textwidth}
	\includegraphics[width=\textwidth]{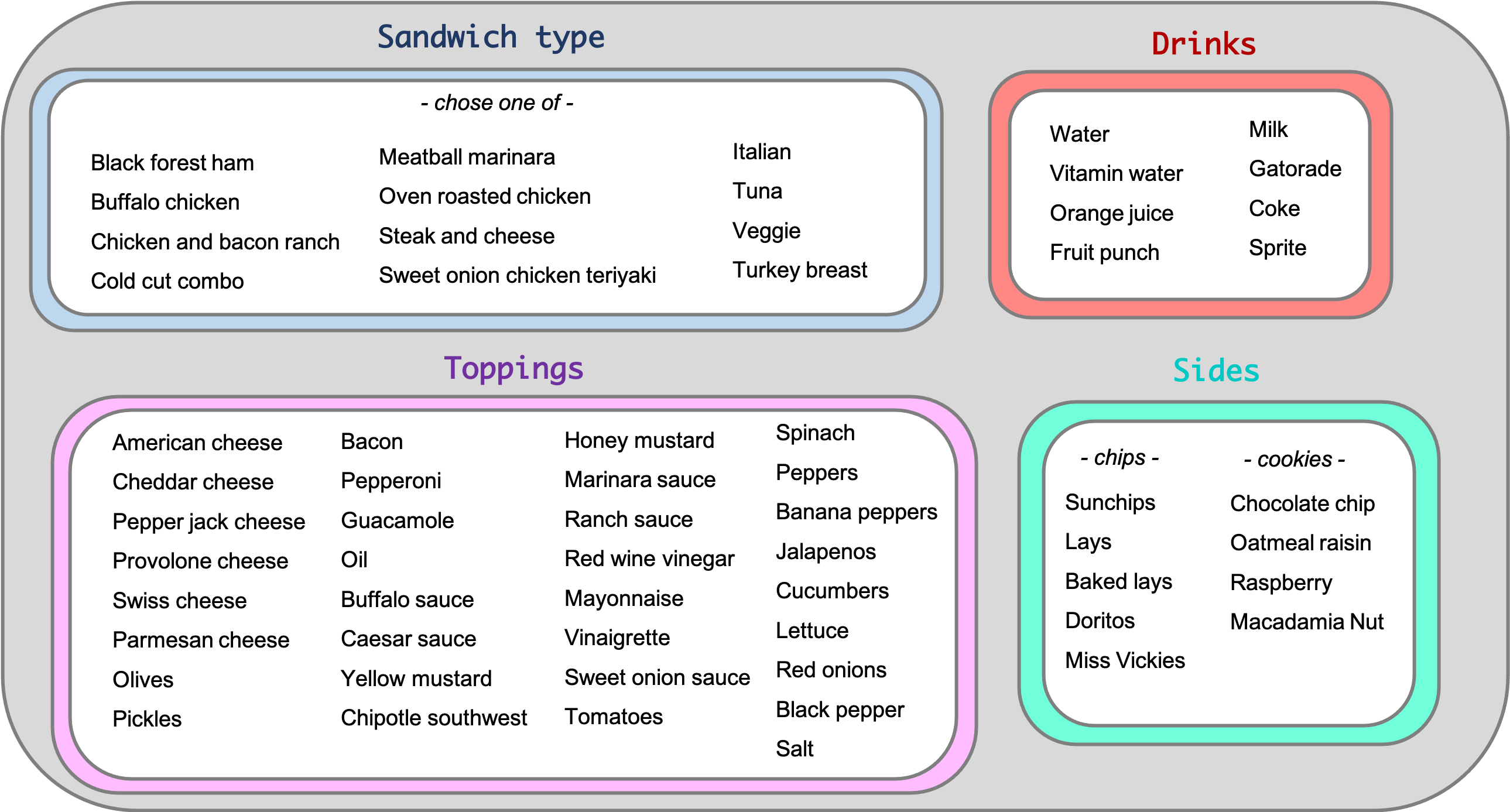}
	\end{subfigure}
	\vskip\baselineskip
	\begin{subfigure}[b]{0.475\textwidth}
	\includegraphics[width=\textwidth]{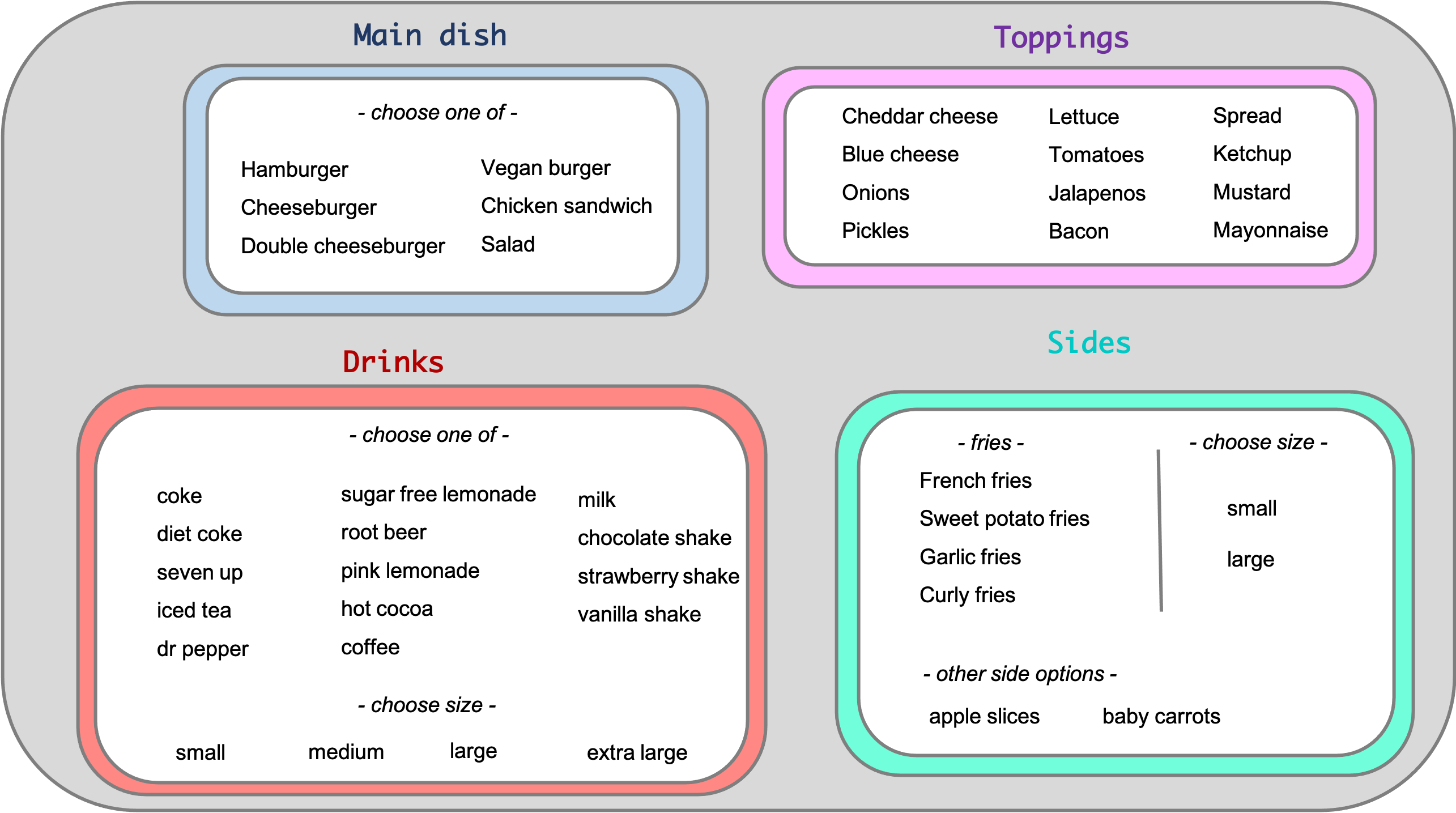}
	\end{subfigure}
	\hfill
	\begin{subfigure}[b]{0.475\textwidth}
	\includegraphics[width=\textwidth]{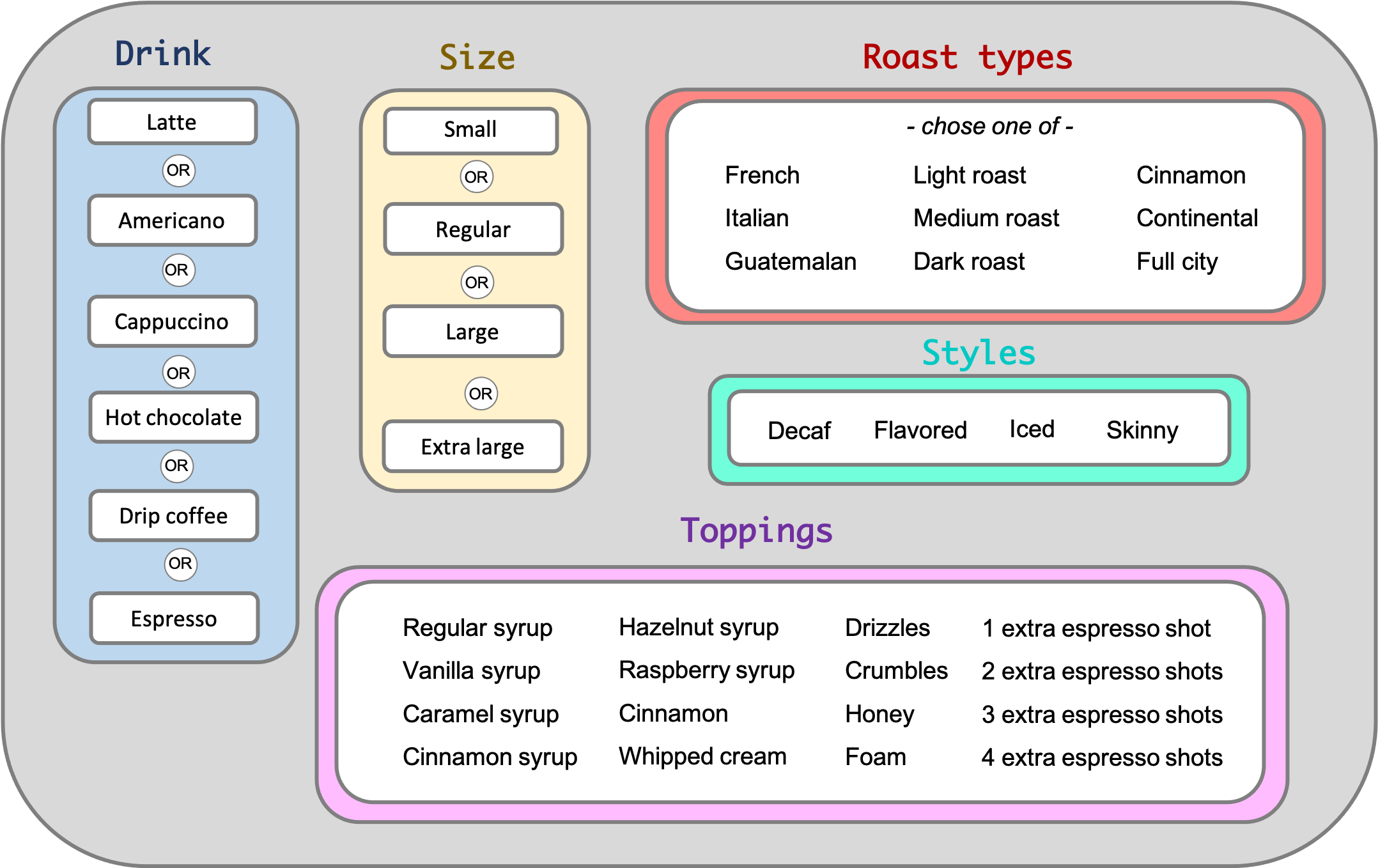}
	\end{subfigure}
	\caption{Menus shown to Mechanical Turk workers for each task: \textsc{burrito} (top-left), \textsc{Sub} (top-right), \textsc{Burger} (bottom-left) and \textsc{coffee} (bottom-right).}\label{fig:allMenus}
\end{figure*}

The instructions given to the workers were templated as shown in Figure \ref{fig:MturkInstructions}. The tasks can be described as natural language text generation with a constrained menu. The number of responses was limited to 3 submissions per worker in order to balance diversity of responses and responsiveness ratios. The respondent's location had to be either US or CA, and the \emph{master} worker qualification was required.\footnote{workers with high ratings according to MTurk API.} The tasks were designed, timed and priced to ensure that the compensation of respondents lies above the US and CA minimum hourly wages. The dataset went through an internal review process to ensure it abides by the company's required standards. Overall we collected answers form about 60 distinct workers for \textsc{burger},  \textsc{sub} and  \textsc{coffee} and about 90 for \textsc{burrito}, for a total of 183 unique individuals.
The menus used for each collection are given in Figure \ref{fig:allMenus}.

\begin{figure*}[b]
	
	\begin{note}
	\scriptsize
	Suppose you want to place your usual order at your favorite \emph{type of restaurant}  (like \emph{examples of such venues}) for you, your partner, your family or your group of friends. 
	Your task is to enter your order exactly as you would say it, verbatim, when you place the order at that restaurant.

	\medskip

	IMPORTANT: This restaurant has a limited menu provided below. Only order items available on the menu, but do so with the same words you usually use when ordering these items:
	
	\medskip
	
	\begin{center}
	 \emph{*** Picture of restaurant Menu ***}
	\end{center}

		\medskip
	
	Write as you would speak. Make sure that:
	\begin{itemize}
	 \item you write your order exactly as you would say it
	 \item your usual order may include many items and if so, include them all when you enter your order below
   	\item if you complete multiple HITs, vary the type of orders you place. The orders should be usual orders you, your friends or family place, but with varying number or types of items, toppings, sides or drinks.
	\end{itemize}

	\medskip
	Enter your order below, using the limited menu above, exactly as you would say it at the restaurant :

	\medskip	
	
	\begin{center}
	 \emph{*** Type order here ***}
	\end{center}

	\end{note}
\caption{Template prompt given to Mechanical Turk workers, common across all 4 tasks. The only significant attribute varying across tasks was the menu to order from.}\label{fig:MturkInstructions}

\end{figure*}

\section{Task Schemas}\label{appendix:skillSchemas}

Detailed schemas for each task,  describing intent names, slot names, and slot properties, are given as supplementary material, along with the full catalog values for each slot.\footnote{\url{https://github.com/amazon-research/food-ordering-semantic-parsing-dataset}}
For illustration purposes, Figure~\ref{fig:allSchema} shows the schema for \textsc{burrito}. Note that not all schemas need to share identical slots for similar intents. For example the \textsc{burger} schema has a \texttt{SIZE} slot for the \texttt{DRINK\_ORDER} and \texttt{SIDE\_ORDER} intents, while the \textsc{burrito} task does not. This is a design choice meant to reflect a real-life setting where the back end for one restaurant might support such property while another might not. This is a challenging---though realistic---obstacle our model needs to overcome.

\medskip

\begin{figure*}[b]
\tiny
	\centering
\begin{minipage}{0.6\textwidth}
	\tiny
	\begin{subfigure}[b]{\textwidth}	
	\begin{lstlisting}[language=json]
{
  "name": "BURRITO",
  "intents":  [
    {"name": "BURRITO_ORDER",
     "invocation_keywords": ["burrito", "burritos"],
      "slots": [ 
      {"name": "NUMBER"}, {"name": "MAIN_FILLING", "quantifiable":  true}, 
      {"name": "RICE_FILLING", "negatable": true, "quantifiable":  true},
      {"name": "BEAN_FILLING", "negatable": true, "quantifiable":  true}, 
      {"name": "SALSA_TOPPING", "negatable": true, "quantifiable":  true},
      {"name": "TOPPING", "negatable": true, "quantifiable":  true}
      ]
    },
    {"name": "BURRITO_BOWL_ORDER",
      "invocation_keywords": ["burrito bowl", "burrito bowls", "bowl", "bowls"],
      "slots": [ 
      {"name": "NUMBER"}, {"name": "MAIN_FILLING", "quantifiable":  true}, 
      {"name": "RICE_FILLING", "negatable": true, "quantifiable":  true},
      {"name": "BEAN_FILLING", "negatable": true, "quantifiable":  true}, 
      {"name": "SALSA_TOPPING", "negatable": true, "quantifiable":  true},
      {"name": "TOPPING", "negatable": true, "quantifiable":  true}
      ]
    },
    {"name": "SALAD_ORDER",
     "invocation_keywords": ["salad", "salads"],
      "slots": [ 
      {"name": "NUMBER"}, {"name": "MAIN_FILLING", "quantifiable":  true}, 
      {"name": "RICE_FILLING", "negatable": true, "quantifiable":  true},
      {"name": "BEAN_FILLING", "negatable": true, "quantifiable":  true}, 
      {"name": "SALSA_TOPPING", "negatable": true, "quantifiable":  true},
      {"name": "TOPPING", "negatable": true, "quantifiable":  true}
      ]
    },
    {"name": "TACO_ORDER",
     "invocation_keywords": ["taco", "tacos"],
      "slots": [ 
      {"name": "NUMBER"}, {"name": "MAIN_FILLING", "quantifiable":  true}, 
      {"name": "RICE_FILLING", "negatable": true, "quantifiable":  true},
      {"name": "BEAN_FILLING", "negatable": true, "quantifiable":  true}, 
      {"name": "SALSA_TOPPING", "negatable": true, "quantifiable":  true},
      {"name": "TOPPING", "negatable": true, "quantifiable":  true}
      ]
    },
    {"name": "QUESADILLA_ORDER",
     "invocation_keywords": ["quesadilla", "quesadillas"],
      "slots": [ 
      {"name": "NUMBER"}, {"name": "MAIN_FILLING", "quantifiable":  true}, 
      {"name": "RICE_FILLING", "negatable": true, "quantifiable":  true},
      {"name": "BEAN_FILLING", "negatable": true, "quantifiable":  true}, 
      {"name": "SALSA_TOPPING", "negatable": true, "quantifiable":  true},
      {"name": "TOPPING", "negatable": true, "quantifiable":  true}
      ]
    },
    {"name": "SIDE_ORDER",
    "invocation_keywords": ["side of chip", "sides of chips"],
      "slots": [
        {"name": "NUMBER"},
        {"name": "SIDE_TYPE"},
        {"name": "SALSA_TOPPING", "negatable": true, "quantifiable":  true}
      ]
    },
    {"name": "DRINK_ORDER",
     "invocation_keywords": ["drink", "drinks"],
      "slots": [
        {"name": "NUMBER"},
        {"name": "DRINK_TYPE"}
      ]
    }
  ]
}
	\end{lstlisting}
\end{subfigure}
\caption{Task schema for the \textsc{burrito} restaurant.}\label{fig:allSchema}
\end{minipage}
\end{figure*}

\section{Dataset Construction Details}\label{appendix:dataExamples}
Part of the dataset comes from the publicly available \textsc{pizza} dataset~\cite{pizzaDataset}. We are following the conditions of use as defined by the license\footnote{\url{https://github.com/amazon-research/pizza-semantic-parsing-dataset/blob/main/LICENSE}} and will release our dataset under the same license.
The collection of new data was done through Mechanical Turk. Respondents were constrained to submit a single utterance for an order containing potentially more than one sub-order. 
Hence, some utterances contained periods and question marks, indicating a sharp separation between two requests. To better reflect the fact that these users would likely have broken their request into multiple ones in an vocal interaction, we split those utterances into pieces. Other punctuation marks like commas, and non-ASCII characters, were simply removed, but utterances were not split around them. Numerical values were spelled out (e.g., \mbox{\emph{2 large cokes}  $\rightarrow$  \emph{two large cokes}}). Finally, utterance text was lower-cased. Annotation was carried out internally by two annotators located in the US. Utterances displaying too much ambiguity for human annotators were removed. In Table \ref{table:exmap} we provide examples of the collected utterances, and their linearized semantics. As can be seen in the table, utterances have varying degrees of complexity, which results in linearized trees of varying depths and widths. 
Synthetic data was generated by sampling human-designed templates, illustrated in Table~\ref{table:synthEx}. For \textsc{sub} we used 32 templates and for \textsc{burrito} we used 46. Some statistics on the degree of compositionality of human and synthetic orders are given in Table \ref{table:uttStats}.

\begin{table*}[!hbt]
  \centering {\scriptsize   
	\begin{tabular}{|l|l||l|}
		\hline
		\textbf{Dataset} & \textbf{Natural Language} & 	\textbf{Semantic representation after entity resolution} \\	\hline
		 \textsc{pizza} & \makecell[l]{five medium pizzas with tomatoes \\ and ham} & \makecell[l]{\smtt{(PIZZAORDER} \\ $\quad$ \smtt{(NUMBER 5 )} \smtt{(SIZE medium )} \\ $\quad$ \smtt{(TOPPING ham )} \smtt{(TOPPING tomatoes ))} } \\ \hline
		\textsc{pizza} &\makecell[l]{i'll have one pie along with pesto and ham  \\ but avoid olives} & \makecell[l]{\smtt{(PIZZAORDER} \\ $\quad$ \smtt{(NOT (TOPPING olives ) )} \\ $\quad$ \smtt{(NUMBER 1 )} \smtt{(TOPPING ham )} \smtt{(TOPPING pesto ))}}  \\  \hline
		\textsc{pizza} &\makecell[l]{i wanted to have two dr peppers three \\ pepsis and a sprite} & \makecell[l]{\smtt{(DRINKORDER} \\ $\quad$ \smtt{(DRINKTYPE dr\_pepper )} \smtt{(NUMBER 2 ))} \\ \smtt{(DRINKORDER} \\ $\quad$ \smtt{(DRINKTYPE pepsi )} \smtt{(NUMBER 3 ))} \\ \smtt{(DRINKORDER} \\ $\quad$ \smtt{(DRINKTYPE sprite )} \smtt{(NUMBER 1 ))} } \\ \hline
		 \textsc{burrito} &\makecell[l]{burrito with steak cheese guacamole sour \\ cream and fresh tomato salsa} & \makecell[l]{\smtt{(BURRITO\_ORDER} \\ $\quad$ \smtt{(NUMBER 1 )} \smtt{(MAIN\_FILLING steak )} \\ $\quad$ \smtt{(TOPPING cheese ) (TOPPING guacamole )} \\ $\quad$ \smtt{(TOPPING sour\_cream )} \\ $\quad$ \smtt{(SALSA\_TOPPING fresh\_tomato\_salsa ) )} } \\ \hline
		\textsc{burrito} &\makecell[l]{i'd also like a bottled water please} & \makecell[l]{\smtt{(DRINK\_ORDER} \\ $\quad$ \smtt{(NUMBER 1 )} \smtt{(DRINK\_TYPE bottled\_water ))}}  \\  \hline
		\textsc{burrito} & \makecell[l]{i'd like a lemonade with a side of chips} & \makecell[l]{\smtt{(DRINK\_ORDER}  \\ $\quad$ \smtt{(NUMBER 1 )} \smtt{(DRINK\_TYPE tractor\_lemonade )} \\ \smtt{(SIDE\_ORDER} \\ $\quad$ \smtt{(NUMBER 1 )} \smtt{(SIDE\_TYPE chips ))} } \\ \hline
		\textsc{sub} & \makecell[l]{steak and cheese sandwich with lettuce \\ cucumbers and olives}  & \makecell[l]{ \smtt{(SANDWICH\_ORDER (NUMBER 1 ) } \\ $\quad$ \smtt{(BASE\_SANDWICH steak\_and\_cheese )} \\ $\quad$ \smtt{(TOPPING lettuce ) (TOPPING cucumbers )} \\ $\quad$ \smtt{(TOPPING black\_olives ) )}} \\ \hline
		\textsc{sub} & \makecell[l]{i will order a chicken and bacon ranch \\ sandwich  and on that please put \\ american cheese chipotle southwest sauce \\  lettuce tomatoes pickles with a side \\ of doritos and two chocolate chip cookies} &  \makecell[l]{ \smtt{(SANDWICH\_ORDER (NUMBER 1 )} \\ $\quad$ \smtt{(BASE\_SANDWICH chicken\_and\_bacon\_ranch ) } \\  $\quad$ \smtt{(TOPPING american\_cheese )} \\ $\quad$ \smtt{(TOPPING chipotle\_southwest )} \\ $\quad$ \smtt{(TOPPING lettuce ) (TOPPING tomatoes )} \\  $\quad$ \smtt{(TOPPING pickles ) )} \\  \smtt{(SIDE\_ORDER (NUMBER 1 )} \\  $\quad$ \smtt{(SIDE\_TYPE doritos\_nacho\_cheese ) )} \\  \smtt{(SIDE\_ORDER (NUMBER 2 )} \\  $\quad$ \smtt{(SIDE\_TYPE chocolate\_chip ) )}} \\ \hline
		\textsc{burger} & \makecell[l]{hi can i have the double cheeseburger \\ with ketchup and onions and french fries \\ on the side} &  \makecell[l]{ \smtt{(MAIN\_DISH\_ORDER (NUMBER 1 ) } \\ $\quad$ \smtt{(MAIN\_DISH\_TYPE double\_cheese\_burger )} \\  $\quad$ \smtt{(TOPPING ketchup ) (TOPPING onion ) )} \\ \smtt{(SIDE\_ORDER (NUMBER 1 )} \\ $\quad$ \smtt{(SIDE\_TYPE french\_fries ) )} } \\ \hline
		\textsc{burger} & \makecell[l]{veggie burger with lettuce and bacon \\  large curly fry and a small iced tea} &  \makecell[l]{ \smtt{(MAIN\_DISH\_ORDER (NUMBER 1 )} \\  $\quad$ \smtt{(MAIN\_DISH\_TYPE vegan\_burger )} \\  $\quad$ \smtt{(TOPPING lettuce ) (TOPPING bacon )  )} \\ \smtt{(SIDE\_ORDER (NUMBER 1 )} \\ $\quad$ \smtt{(SIZE large ) (SIDE\_TYPE curly\_fries ) )} \\ \smtt{(DRINK\_ORDER (NUMBER 1 )} \\ $\quad$ \smtt{(SIZE small ) (DRINK\_TYPE iced\_tea ) )}} \\ \hline
		\textsc{coffee} & \makecell[l]{i'd like a large hot chocolate with \\ whipped cream} &   \makecell[l]{ \smtt{(DRINK\_ORDER} \\  $\quad$ \smtt{(NUMBER 1 ) (SIZE large )} \\ $\quad$ \smtt{(DRINK\_TYPE hot\_chocolate )} \\  $\quad$ \smtt{(TOPPING whipped\_cream ) )}} \\ \hline
		\textsc{coffee} & \makecell[l]{one regular latte light roast with an \\ extra espresso shot and honey added and \\ one large cappuccino with caramel syrup \\ in that one} & \makecell[l]{ \smtt{(DRINK\_ORDER (NUMBER 1 )} \\ $\quad$ \smtt{(SIZE regular ) (DRINK\_TYPE latte )} \\ $\quad$ \smtt{(ROAST\_TYPE light\_roast ) (TOPPING honey )} \\ $\quad$ \smtt{(TOPPING (ESPRESSO\_SHOT 1 ) ) )} \\ \smtt{(DRINK\_ORDER (NUMBER 1 )} \\ $\quad$ \smtt{(SIZE large ) (DRINK\_TYPE cappuccino )} \\ $\quad$ \smtt{(TOPPING caramel\_syrup ) )} } \\ \hline
		
	\end{tabular}
}        
	\caption{Example utterances obtained from Mechanical Turk collection and their corresponding machine-executable representation.} \label{table:exmap}
\end{table*}

\begin{table*}[!hbt]
  \centering {\scriptsize   
	\begin{tabular}{|l|l||l|}
		\hline
		\textbf{Dataset} & \textbf{Template} & 	\textbf{Example catalog values} \\	\hline
		\textsc{sub} & \makecell[l]{ \smtt{ \{prelude\} \{number\} \{side\_type\} } }   &  \makecell[l]{ \smtt{\{prelude\}} = \textit{i want to order} \\  \smtt{\{side\_type\} } = \textit{sunchips} }  \\ \hline
		\textsc{sub} & \makecell[l]{  \smtt{ \{prelude\} \{number\} \{base\_sandwich\} with} \\  \smtt{\{topping1\} and \{topping2\} } } &  \makecell[l]{ \smtt{\{base\_sandwich\}} = \textit{chicken teriyaki} \\  \smtt{\{topping1\} } = \textit{bacon} } \\ \hline
		\textsc{burrito} & \makecell[l]{ \smtt{ \{prelude\} \{number\} \{main\_filling\} \{entity\_name\} } \\ \smtt{ with \{salsa\_topping\} }  } &   \makecell[l]{ \smtt{\{main\_filling\}} = \textit{barbacoa} \\  \smtt{\{entity\_name\} } = \textit{burrito} } \\ \hline
		\textsc{burrito} & \makecell[l]{ \smtt{\{prelude\} \{number\} side of \{side\_type1\} and} \\ \smtt{ \{side\_type2\} and \{number\} \{drink\_type\} } } & \makecell[l]{ \smtt{\{side\_type1\}} = \textit{chips} \\  \smtt{\{side\_type2\} } = \textit{guac} } \\ \hline
	\end{tabular}
}        
	\caption{Example templates and catalog values used for sampling synthetic data.} \label{table:synthEx}
\end{table*}

\section{Constrained Decoding Details}\label{appendix:constrained}
In what follows we list the actual constraints implemented in this work in the form of allowed transitions. Any element on the left of the arrow can be followed by elements on the right:

\small
\[
\left\{
\begin{tabular}{lll}
\texttt{BOS} & $\rightarrow$ & \{$"( X "$,  $X$ = valid intent \} \\
$"("$ & $\rightarrow$ &  \{ $X$,  $X$ = valid intent or slot\} \\
$")"$ & $\rightarrow$ & \{$")"$ or $"("$ or \texttt{EOS} \} \\
intent  & $\rightarrow$   & \{$"( X "$, $X$ = a valid slot  \} \\
slot & $\rightarrow$   & \{$X$,  $X$ = compatible value \} \\
\texttt{(COMPLEX}   & $\rightarrow$ &\texttt{(QUANTITY}\
\end{tabular}
\right.
\]

\normalsize
 One could think of imposing more grammar-based constraints, for example, allowing only valid intent-slot combinations, or  only allowing negatable slots after \texttt{(NOT}, since some of these---like \texttt{SIZE}---cannot be negated. Examples of how constrained decoding helped can be found in Table~\ref{table:constrained_decoding_helps_examples}.

 \begin{center}
 	\begin{table*}[hbt!]
 			\centering
 			\begin{center}
 			\small
 				\begin{tabular}[c]{lccc}
 					\midrule
 
 					 				& \makecell[c]{\#Intent per \\ utterance}  & \makecell[c]{\#Slots per \\ utterance} & \makecell[c]{Avg utterance \\ depth}    \\ 
 					 \midrule
 					 Synthetic Data	& & &											\\
 					\midrule
 					\textsc{pizza}                      & 1.77	&	5.77	&	3.44 \\
 					\textsc{burrito}                      & 1.57	&	6.50	&	3.48\\
 					\textsc{sub}                       & 1.79	&	6.24	&	3.37 \\
 					\midrule
 					 Human-generated Data	& & &											\\
  					\midrule
 					\textsc{pizza}                      & 1.25	&	6.13	&	3.62  \\
 					\textsc{burrito}                      & 1.39	&	5.78	&	3.12 \\
 					\textsc{sub}                        & 1.69	&	5.99	&	3.07\\
 					\textsc{burger}                      & 1.97	&	7.17	&	3.04 \\
 					\textsc{coffee}                      & 1.05	&	5.34	&	3.2 \\
 					\hline
 				\end{tabular}
 			\end{center}
 			\caption{Statistics on the degree of compositionality in each task, for synthetic and human-generated data.}
 			\label{table:uttStats}
 	\end{table*}
 \end{center}
 
\begin{table*}[!hbt]
  \hspace*{-1cm}{\footnotesize   
	\begin{tabular}{|l|l||l|l|}
		\hline
		\textbf{Dataset} & \textbf{Natural Language Utterance} & 	\textbf{Prediction w/o constraints} & 	\textbf{Prediction w/ constraints} \\	\hline
		 \textsc{burger} &\makecell[l]{i'll have a hamburger topped with \\ bacon and ketchup along with a large \\ coke and large order of french fries} & \makecell[l]{\smtt{(MAIN\_DISH\_ORDER} \\ $\quad$ \smtt{(MAIN\_DISH\_TYPE hamburger )} \\ $\quad$ \smtt{(TOPPING bacon )} \\ $\quad$ \smtt{(TOPPING ketchup ))} \\ \smtt{(DRINK\_ORDER} \\ $\quad$\smtt{(SIZE large )} \\ $\quad$ \smtt{(DRINK\_TYPE coke ))} \\ \smtt{(SIDE\_ORDER} \smtt{(\textcolor{red}{NUMBER large} )} \\ $\quad$\smtt{(SIDE\_TYPE french fries ))} } & \makecell[l]{\smtt{(MAIN\_DISH\_ORDER} \\ $\quad$ \smtt{(MAIN\_DISH\_TYPE hamburger )} \\ $\quad$ \smtt{(TOPPING bacon )} \\ $\quad$ \smtt{(TOPPING ketchup ))} \\ \smtt{(DRINK\_ORDER} \\ $\quad$\smtt{(SIZE large )} \\ $\quad$ \smtt{(DRINK\_TYPE coke ))} \\ \smtt{(SIDE\_ORDER} \smtt{(\textcolor{green}{NUMBER a} )} \\ $\quad$ \smtt{(\textcolor{green}{SIZE large} )} \\ $\quad$ \smtt{(SIDE\_TYPE french fries ))} } \\ \hline
		 \textsc{coffee} &\makecell[l]{i'd like an iced cappuccino with \\ caramel syrup and whipped cream} & \makecell[l]{\smtt{(DRINK\_ORDER} \\ $\quad$ \smtt{(\textcolor{red}{STYLE iced cappuccino} )} \\ $\quad$ \smtt{(TOPPING caramel syrup  )} \\ $\quad$ \smtt{(TOPPING whipped cream ))} } & \makecell[l]{\smtt{(DRINK\_ORDER} \\ $\quad$ \smtt{(\textcolor{green}{STYLE iced} )} \\ $\quad$ \smtt{(\textcolor{green}{DRINK\_TYPE cappuccino} )} \\ $\quad$ \smtt{(TOPPING caramel syrup )} \\ $\quad$ \smtt{(TOPPING whipped cream ))} } \\ \hline		
	\end{tabular}
}
	\caption{Example utterances where constrained decoding helps fix invalid slot/slot value combinations.} 
    \label{table:constrained_decoding_helps_examples}
\end{table*}

\section{Computational Details}\label{appendix:comput}
Hyperparameter tuning was performed on learning rates [\texttt{5e-04}, \texttt{1e-05}, \texttt{5e-05}, \texttt{1e-06}]
and batch sizes [\texttt{16, 24, 48, 64}] across three seeds. We use the human-generated data of the three training tasks as our development set for early stopping and hyperparameter tuning. Including this and general experimentation, we estimate our total computation cost to be about 2 weeks GPU hours.

\end{document}